\documentclass[a4paper]{article}

\usepackage[utf8]{inputenc}
\usepackage{erk}
\usepackage{times}
\usepackage{graphicx}
\usepackage{flafter}
\usepackage{booktabs}
\usepackage[top=22.5mm, bottom=22.5mm, left=22.5mm, right=22.5mm]{geometry}
\usepackage[slovene,english]{babel}

\def\footnotemark{}
\begin{document}
%make title
\title{Evaluation of Monocular SLAM Systems on High-Altitude Nadir UAV Footage}

\author{Gašper Spagnolo$^{1}$, Matej Dobrevski$^{1}$, Danijel Skočaj$^{1}$} 

\affiliation{$^{1}$ Faculty of Computer and Information Science, University of Ljubljana, Večna pot 113, Ljubljana, Slovenia}

\email{E-pošta: spagnolo.gasper@gmail.com}

\maketitle

\begin{abstract}{Abstract}
Aerial nadir video combines weak geometric constraints with severe perceptual aliasing, making it a difficult regime for monocular SLAM. We benchmark five monocular SLAM systems on local UAV flights, synthetic city-scale imagery, and long-range aerial sequences. To isolate visual performance, we provide no inertial or GNSS aiding. Performance varies strongly with environment and trajectory scale: MASt3R-SLAM achieves the lowest mean horizontal MAE on the five DJI flights (0.53\% of reference path length), whereas no system consistently preserves global trajectory shape on the long GES and ALTO sequences. Overall, DROID-SLAM performs best, averaging 2.88\% of reference path length across completed runs. Vertical position remains poor, and large-area trajectories remain highly distorted despite loop-closure capability. Current monocular SLAM methods are by themselves therefore insufficient for reliable visual-only aerial navigation.
\end{abstract}

\selectlanguage{english}

\section{Introduction}

Unmanned aerial vehicles (UAVs) depend heavily on Global Navigation Satellite Systems (GNSS) for localization despite their vulnerability to jamming, spoofing, and signal denial. Visual odometry (VO), which estimates relative camera motion from consecutive images, offers a camera-based fallback.

\begin{figure}[!t]
    \begin{center}
        \includegraphics[width=\linewidth]{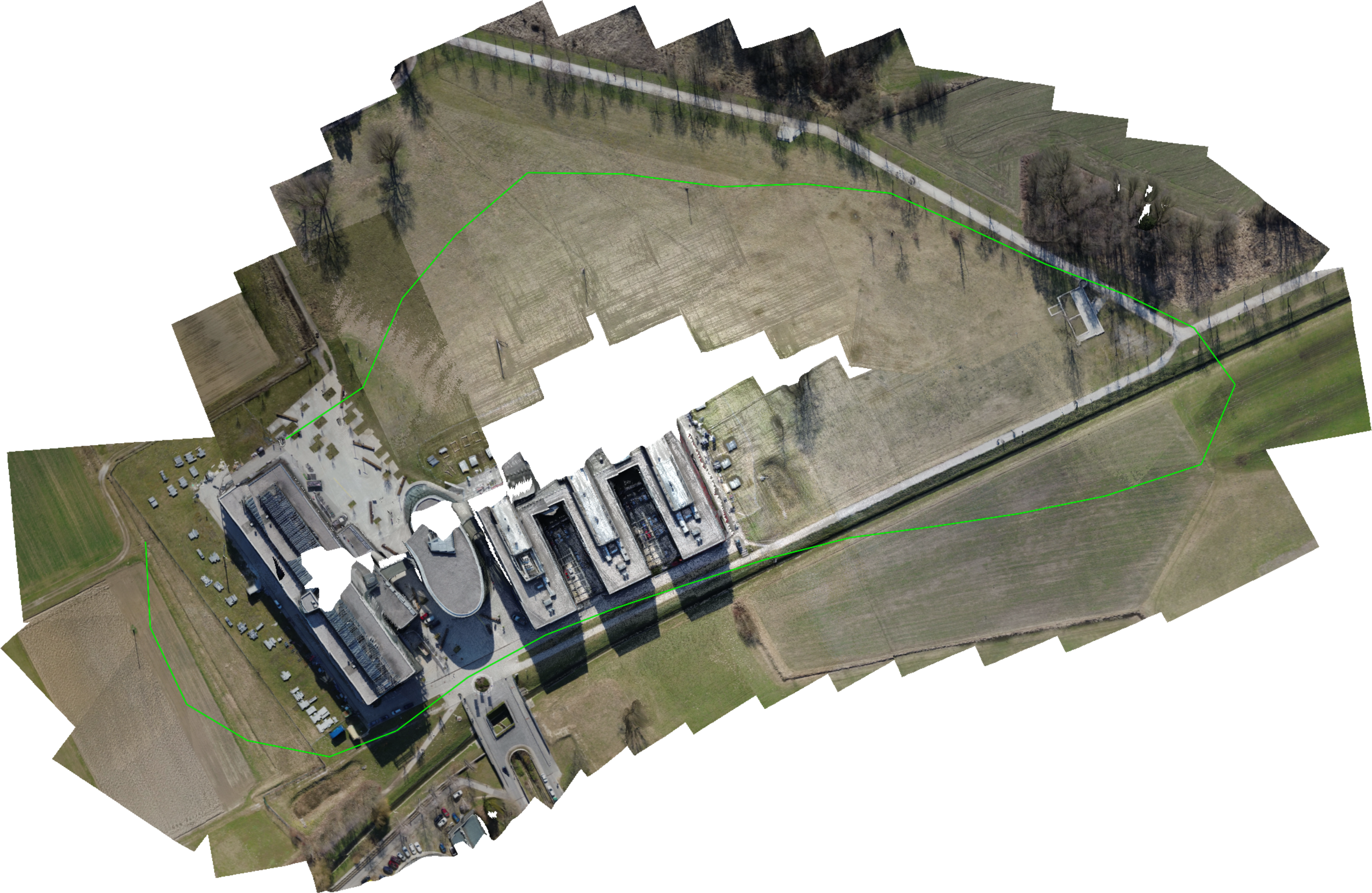}
        \caption{FRI sequence of our DJI dataset: nadir frames composited over the flight area, with the flight trajectory shown in green.} \label{fig:dji-seq}
    \end{center}
\end{figure}

VO alone cannot provide long-term consistency because errors compound from one pose estimate to the next. Simultaneous localization and mapping (SLAM) augments sequential tracking with a global map, typically optimizing a graph whose nodes are camera poses and whose edges encode tracked motion or loop closures. Graph optimization can distribute accumulated drift after a revisit, while relocalization can resume tracking after a failure. Both depend on reliable geometry and place recognition.

Modern visual SLAM is usually developed and benchmarked on indoor, ground-level, or low-altitude forward-facing video. High-altitude nadir views provide little translational parallax, scene-depth variation is small relative to the viewing distance, and terrain can be weakly textured or highly repetitive. The resulting near-planar geometry is degenerate for two-view pose estimation~\cite{spagnolo2024erk}. These conditions also undermine loop closure: long survey paths contain few revisits, while similar-looking terrain makes genuine revisits hard to distinguish from aliases. It is therefore unclear whether global SLAM optimization actually limits drift in this setting.

To answer this question, we establish a common protocol to compare SOTA visual SLAM approaches, both deep-learning and conventional: ORB-SLAM3~\cite{campos2021orbslam3}, DROID-SLAM~\cite{teed2021droidslam}, DPVO~\cite{teed2023dpvo}, MASt3R-SLAM~\cite{murai2025mast3rslam}, and VGGT-SLAM~2.0~\cite{maggio2026vggtslam2}. We measure trajectory error and inspect tracking and loop-closure failures on consumer DJI flights (Figure~\ref{fig:dji-seq}), city-scale Google Earth Studio renders~\cite{ges}, and the public ALTO~\cite{cisneros2022alto} and MARS-LVIG~\cite{li2024marslvig} datasets.

\section{Related Work}

Classical visual SLAM couples sparse image features with geometric optimization. PTAM~\cite{klein2007ptam} established keyframe-based parallel tracking and mapping, and the ORB-SLAM series~\cite{murartal2015orbslam,campos2021orbslam3} made this architecture broadly competitive. Reliance on repeatable local features leaves such pipelines sensitive to motion blur, low texture, and rotation-dominant motion, all frequent in aerial footage.

Learned SLAM systems replace hand-crafted correspondence or update rules with trained networks~\cite{teed2021droidslam,teed2023dpvo,lipson2024dpvslam}. Like the earlier TartanVO model~\cite{wang2021tartanvo}, these systems are commonly trained on synthetic TartanAir~\cite{wang2020tartanair}, whose aggressive drone-like trajectories make them natural candidates for aerial deployment.

3D foundation models instead infer scene geometry directly from images. DUSt3R~\cite{wang2024dust3r} introduced pairwise pointmap regression, MASt3R~\cite{leroy2024mast3r} added dense matching features, and VGGT~\cite{wang2025vggt} extended joint camera and geometry prediction to many views. Related work adds spatial memory, persistent state, and more efficient multi-view processing~\cite{wang2025spann3r,yang2025fast3r,wang2025cut3r,liu2025slam3r}. These priors now support sequential systems including MASt3R-SLAM, VGGT-SLAM and its 2.0 revision, ViSTA-SLAM, and VGGT-Long~\cite{murai2025mast3rslam,maggio2025vggtslam,maggio2026vggtslam2,zhang2026vistaslam,deng2026vggtlong}.

Aerial evaluations of SLAM remain scarce and low-altitude: established UAV benchmarks cover indoor flight (EuRoC~\cite{burri2016euroc}), racing (UZH-FPV~\cite{delmerico2019uzhfpv}), and synthetic low-altitude flight (Mid-Air~\cite{fonder2019midair}, TartanAir~\cite{wang2020tartanair}). Higher altitudes appeared only recently, through the nadir MARS-LVIG dataset~\cite{li2024marslvig}, a visual-inertial odometry study at 40-100\,m~\cite{george2023vio}, and a nadir-view fusion system showing that forward-facing formulations degrade on such data~\cite{li2025nadir}. Closest to our setting, pointmap models were evaluated on photogrammetric aerial blocks~\cite{wu2025aerial}, while aerial-domain fine-tuning improved pose estimation under extreme ground-aerial viewpoint changes~\cite{vuong2025aerialmegadepth}. Both concern offline reconstruction rather than sequential SLAM. Our previous work~\cite{spagnolo2024erk} benchmarked image matchers within UAV visual odometry. A systematic evaluation of complete SLAM systems on high-altitude nadir footage is, to our knowledge, still missing.

\section{Methods}

We compare five monocular SLAM systems spanning feature-based, recurrent, and foundation-model approaches.

\textbf{ORB-SLAM3}~\cite{campos2021orbslam3} combines ORB features, local bundle adjustment over a covisibility graph, and bag-of-words place recognition. Its Atlas creates a new map after tracking loss and merges maps on revisits. For fragmented runs, we evaluate the largest map.

\textbf{DROID-SLAM}~\cite{teed2021droidslam} maintains a dynamic graph of co-visible keyframes. A recurrent operator refines dense correspondences and confidence weights, while differentiable bundle adjustment updates depth and pose. A backend periodically optimizes the full keyframe history and provides loop closure.

\textbf{DPVO}~\cite{teed2023dpvo} maintains a sparse patch graph over a sliding window. Its recurrent update operator predicts patch-trajectory corrections and confidence weights, while differentiable bundle adjustment optimizes camera poses and patch inverse depths. We treat it as local SLAM because it jointly estimates motion and a local scene representation, although it lacks loop closure and full-history global optimization.

\textbf{MASt3R-SLAM}~\cite{murai2025mast3rslam} uses a frozen two-view model~\cite{leroy2024mast3r} to predict per-pixel pointmaps and dense descriptors. Tracking minimizes a confidence-weighted ray error against the keyframe's fused pointmap. An ASMK database~\cite{tolias2013asmk,tolias2020deep} aggregates MASt3R local descriptors for loop-candidate retrieval, and a second-order optimization of the keyframe graph maintains global consistency.

\textbf{VGGT-SLAM 2.0}~\cite{maggio2026vggtslam2} reconstructs overlapping submaps with VGGT~\cite{wang2025vggt} and optimizes an $\mathrm{SL}(4)$ keyframe graph whose intra-submap factors encode rotation and translation and whose inter-submap factors encode calibration and scale. SALAD global descriptors~\cite{izquierdo2024salad} propose loop candidates, which a score derived from VGGT's layer-22 attention verifies. This constrained factorization removes the original system's unconstrained 15-DoF drift and planar degeneracy~\cite{maggio2025vggtslam}.

\section{Datasets}
\label{sec:datasets}

Four diverse aerial datasets span local UAV surveys to city-scale trajectories. All methods use monocular RGB only.

\textbf{DJI.} Five short nadir sequences capture comparatively lower-altitude flights over rural and campus environments. They were recorded at 110-120\,m above ground around the Faculty of Computer and Information Science in Ljubljana and in Šempeter pri Gorici with a DJI Mini 3, a consumer quadrotor with a gimbal-stabilized camera ($82.1^\circ$ field of view)~\cite{djimini3}. The reference trajectory is the onboard multi-GNSS position embedded in the video metadata at 1\,Hz. Without RTK corrections, its accuracy is limited.

\textbf{GES.} Three long, higher-altitude (150-800\,m), predominantly urban synthetic trajectories were generated with Google Earth Studio~\cite{ges}, which exports camera poses and intrinsics consistent with the rendered sequence. They cover Ljubljana (83.77\,km ground path), Graz (57.21\,km), and Venice (81.28\,km), rendered with an $80^\circ$ vertical field of view and totaling 222\,km of ground-track motion. The setup follows our previous evaluation~\cite{spagnolo2024erk}. Compressing each path into 100\,s of 30\,Hz footage makes GES a long-range visual stress test rather than a dynamically realistic UAV flight.

\textbf{ALTO}~\cite{cisneros2022alto} contains long, higher-altitude helicopter flights over largely rural and vegetated terrain in Ohio and Pennsylvania, recorded with a nadir global-shutter RGB camera and high-precision GPS--INS reference poses. The full dataset is not public, so we use two nested GPR Competition releases~\cite{metaslamgpr}: their training splits cover 28.5 and 37.4\,km, with the shorter a prefix of the longer, and both share the same 4.6\,km validation stream. These entries are therefore not independent.

\textbf{MARS-LVIG}~\cite{li2024marslvig} provides short, comparatively lower-altitude flights across varied airfield, island, town, and valley environments. The nadir multi-sensor UAV sequences were flown at 80-130\,m altitude and 3-12\,m/s, with ground truth from the RTK receiver of the carrier UAV. We use only its RGB image stream, trimmed of takeoff and landing so that all sequences contain cruise flight only.

\begin{figure*}[!t]
    \begin{center}
        \includegraphics[width=0.95\textwidth]{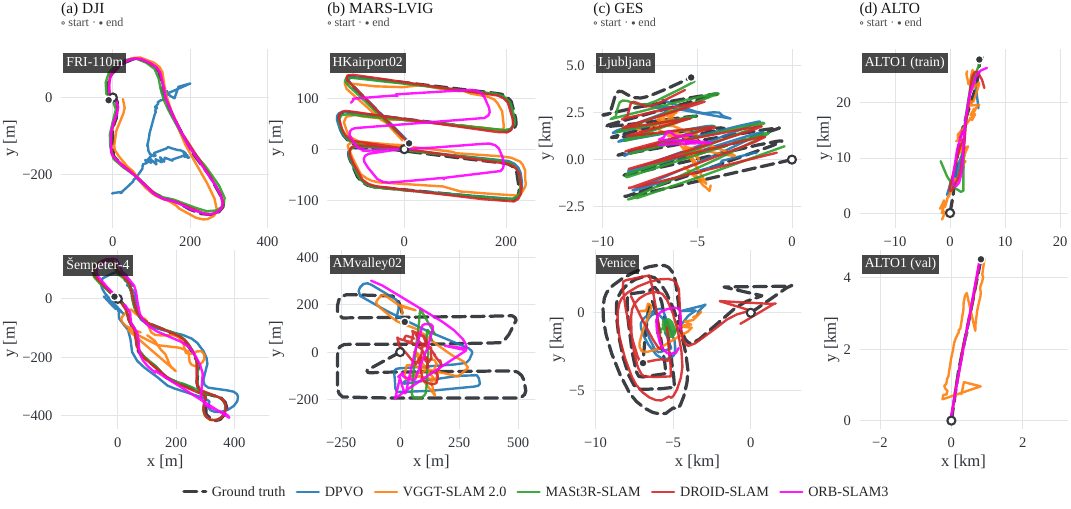}
        \caption{Representative aligned trajectories for two contrasting sequences from each dataset. Note the different axis scales: DJI and MARS-LVIG sequences span hundreds of meters, whereas GES and ALTO span several kilometers.} \label{fig:trajs}
    \end{center}
\end{figure*}

\subsection{Evaluation protocol}

Because the systems return poses at different densities, from every frame to sparse keyframes, and do not preserve timestamps consistently, index-wise association would pair unrelated poses. We instead interpolate both trajectories over normalized horizontal progress (0-100\%) and align them with a planar similarity transform (XY rotation and translation plus global scale) using Umeyama~\cite{umeyama1991}. Both tables report relative horizontal MAE,
\[
E_{\mathrm{rel}}(\%) = \frac{100}{N L_{\mathrm{ref}}}
\sum_{i=1}^{N}\left\|\hat{\mathbf p}^{\,a}_{i,xy}-\mathbf p_{i,xy}\right\|_2,
\]
where $N$ is the number of comparisons, $L_{\mathrm{ref}}$ is the reference XY path length, and $\mathbf p_{i,xy}$ and $\hat{\mathbf p}^{\,a}_{i,xy}$ are the reference and aligned-estimate positions. Unlike timestamp-associated ATE~\cite{sturm2012benchmark}, this dimensionless path-shape error supports comparison across sequence lengths. Table~\ref{tab:ate-seq} lists individual sequences. Table~\ref{tab:ate-agg} gives mean $\pm$ sample SD over completed runs. Vertical error uses only the fitted scale, without vertical rotation or translation. Unevaluable runs are excluded. All experiments use an NVIDIA A100 (40\,GB) GPU.

\section{Results}

Tables~\ref{tab:ate-seq} and~\ref{tab:ate-agg} summarize horizontal error, while Figure~\ref{fig:trajs} shows representative aligned trajectories and failure modes. The results are given as the Mean Absolute Error, divided by the total length of the trajectory. In this way, we get an error as percentage of the traveled distance. We omit vertical values from the tables because vertical RMSE exceeds 100\,m in 29 of 34 completed GES and ALTO runs, including several with low horizontal error.

\begin{table}[!htb]
\caption{Per-sequence horizontal MAE divided by reference path length (\%). The best result in each row is shown in bold. ``--'' marks a failed run.} \label{tab:ate-seq}
\smallskip
\centering
\footnotesize
\setlength{\tabcolsep}{3pt}
\begin{tabular}{ l r r r r r }
\toprule
\textbf{Sequence} & \textbf{DPVO} & \textbf{DROID} & \textbf{MASt3R} & \textbf{ORB3} & \textbf{VGGT2} \\
\midrule
\multicolumn{6}{l}{\textit{DJI}} \\
FRI-110m & 10.47 & \textbf{0.16} & 0.70 & 0.22 & 1.09 \\
Šempeter-1 & 4.54 & \textbf{0.20} & 1.03 & 2.25 & 0.39 \\
Šempeter-2 & 8.71 & 1.92 & \textbf{0.34} & 7.74 & 8.38 \\
Šempeter-3 & 1.96 & 0.25 & 0.29 & \textbf{0.21} & 11.13 \\
Šempeter-4 & 3.09 & 0.37 & \textbf{0.29} & 1.57 & 10.74 \\
\midrule
\multicolumn{6}{l}{\textit{GES}} \\
Ljubljana & 1.53 & 1.69 & \textbf{1.17} & 2.88 & 2.58 \\
Venice & 3.97 & \textbf{2.05} & 4.16 & 4.08 & 3.90 \\
Graz & \textbf{0.05} & 1.73 & 5.78 & 5.65 & 5.18 \\
\midrule
\multicolumn{6}{l}{\textit{MARS-LVIG}} \\
AMtown01 & 1.33 & 0.86 & 3.59 & \textbf{0.34} & 1.84 \\
AMtown02 & 2.78 & 2.11 & \textbf{2.06} & 4.95 & 3.42 \\
AMtown03 & 3.62 & 2.82 & \textbf{0.58} & 5.33 & 4.25 \\
AMvalley01 & 1.29 & 2.25 & 5.28 & \textbf{0.96} & 5.46 \\
AMvalley02 & \textbf{4.23} & 6.37 & 5.87 & 5.25 & 5.31 \\
AMvalley03 & 5.77 & 5.12 & 6.47 & 6.31 & \textbf{3.74} \\
HKairport01 & \textbf{0.07} & 0.24 & 6.22 & 0.65 & 1.04 \\
HKairport02 & 0.51 & \textbf{0.26} & 0.45 & 3.66 & 1.70 \\
HKairport03 & 1.46 & \textbf{0.43} & 6.03 & 3.17 & 0.92 \\
HKisland01 & 0.60 & 0.68 & 4.92 & \textbf{0.08} & 1.74 \\
HKisland02 & 1.55 & \textbf{1.09} & 5.03 & 1.16 & 3.52 \\
HKisland03 & \textbf{3.67} & 6.35 & -- & 5.24 & 4.05 \\
\midrule
\multicolumn{6}{l}{\textit{ALTO}} \\
ALTO1 (train) & 10.48 & 10.99 & 6.70 & 13.05 & \textbf{4.69} \\
ALTO1 (val) & \textbf{0.19} & 0.22 & 23.84 & 0.31 & 10.42 \\
ALTO2 (train) & 23.42 & 20.69 & -- & 12.80 & \textbf{7.17} \\
ALTO2 (val) & \textbf{0.20} & 0.22 & 24.55 & 0.30 & 10.42 \\
\bottomrule
\end{tabular}
\end{table}

\begin{figure*}[!t]
    \centering
    \includegraphics[width=0.95\textwidth]{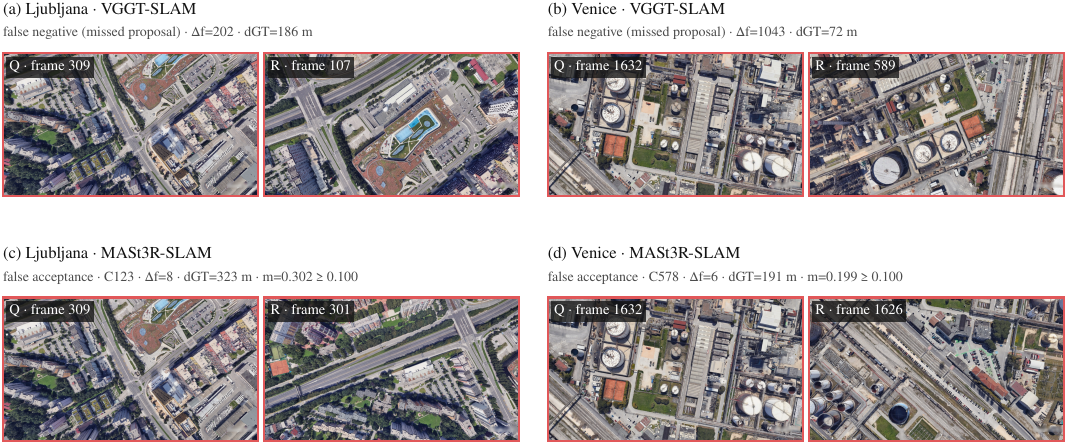}
    \caption{Representative loop-closure failures on GES. In (a,b), ground truth and image overlap confirm true revisits that VGGT-SLAM~2.0 never proposes for attention verification. In (c,d), MASt3R-SLAM accepts spatially inconsistent pairs. Both are sequential candidates, as indicated by their small frame-index gaps. $Q$ and $R$ denote query and retrieved frames, $\Delta f$ their frame-index separation, and $d_{\mathrm{GT}}$ their ground-truth distance. For MASt3R-SLAM, $C$ is the candidate identifier and $m$ the geometric match fraction relative to its threshold. Red borders mark erroneous outcomes.} \label{fig:loop-failures}
\end{figure*}

\begin{table}[!htb]
\caption{Horizontal MAE divided by reference path length (\%), reported as mean $\pm$ sample SD over completed runs.}
\label{tab:ate-agg}
\smallskip
\centering
\scriptsize
\setlength{\tabcolsep}{1.5pt}
\begin{tabular}{l c c c c c}
\toprule
& \textbf{DPVO} & \textbf{DROID} & \textbf{MASt3R} & \textbf{ORB3} & \textbf{VGGT2} \\
\midrule
DJI & 5.75$\pm$3.67 & 0.58$\pm$0.75 & \textbf{0.53$\pm$0.33} & 2.40$\pm$3.11 & 6.34$\pm$5.23 \\
GES & 1.85$\pm$1.98 & \textbf{1.82$\pm$0.20} & 3.71$\pm$2.34 & 4.20$\pm$1.39 & 3.89$\pm$1.30 \\
MARS-LVIG & \textbf{2.24$\pm$1.75} & 2.38$\pm$2.32 & 4.23$\pm$2.23 & 3.09$\pm$2.32 & 3.08$\pm$1.59 \\
ALTO & 8.57$\pm$11.02 & 8.03$\pm$9.85 & 18.36$\pm$10.11 & \textbf{6.61$\pm$7.29} & 8.18$\pm$2.79 \\
\midrule
\textbf{ALL} & 3.98$\pm$5.15 & \textbf{2.88$\pm$4.62} & 5.24$\pm$6.59 & 3.67$\pm$3.66 & 4.71$\pm$3.36 \\
\bottomrule
\end{tabular}
\end{table}

On the real DJI sequences, MASt3R-SLAM is the most accurate and consistent, with DROID-SLAM close behind. DPVO is unstable throughout, while VGGT-SLAM ranges from second-best to worst.

On MARS-LVIG, rankings cluster more tightly: DPVO and DROID-SLAM lead on the airfield sequences, the valleys are the hardest setting, and the winner changes across environments.

At city scale the picture inverts. Except for DPVO on Graz (31.0\,m), GES errors range from 1.11 to 3.84\,km. On the long ALTO training splits, no method yields a usable trajectory.

Failure modes are system-specific (Figure~\ref{fig:trajs}). ORB-SLAM3 repeatedly loses local tracking on Šempeter-2, yet its ALTO runs retain a single Atlas map despite severe trajectory distortion: at survey scale, failure can appear as progressive deformation rather than an explicit loss. DROID-SLAM likewise drifts progressively on large-area sequences without a discrete failure. DPVO exhibits one-way drift, with estimates leaving the survey area while the true flight returns upon itself. MASt3R-SLAM performs best on the built-up DJI sequences and poorly over homogeneous valleys, suggesting sensitivity to scene content and depth variation.

The loop-closure pipelines show opposite biases (Figure~\ref{fig:loop-failures}): VGGT-SLAM appears overly conservative, with SALAD missing genuine revisits and attention verification rejecting most proposals, whereas MASt3R-SLAM appears overly permissive, accepting many ASMK-retrieved pairs that are temporally local and some that are spatially inconsistent. Across all three GES sequences, VGGT-SLAM proposes 71 candidates over 96 submaps, of which attention-based retrieval verification rejects 66 (93\%). Ground-truth inspection also identifies missed proposals: in Figure~\ref{fig:loop-failures}(a,b), the query and earlier frame revisit the same place, but SALAD never proposes the pair, so the verifier cannot recover it. Despite five accepted constraints, VGGT-SLAM retains 2.4-3.5\,km horizontal normalized-path error. On Ljubljana, MASt3R-SLAM accepts 468 of 1016 candidates, but most are temporally local: 320 connect keyframes two positions apart, and 386 (82\%) lie within five keyframes. The accepted set also includes candidates with large ground-truth separation. The Ljubljana and Venice examples in Figure~\ref{fig:loop-failures}(c,d) are only eight and six frames apart but 323\,m and 191\,m apart in ground truth. One plausible contributor is the retrieval representation. ASMK scores selective matches between aggregated local descriptors, whereas SALAD represents each image globally. This difference may preserve partial overlap in ASMK while making repeated structures ambiguous, whereas global pooling can miss partial overlap. However, the present runs do not isolate representation from thresholds or verification. In neither case does loop closure arrest large-area drift.

Resource demand further narrows the practical deployment options. ORB-SLAM3 is the only evaluated method that runs entirely on a CPU. DPVO is the most viable learned alternative when a GPU is available because it uses less than 3\,GB of VRAM. In contrast, DROID-SLAM, MASt3R-SLAM, and VGGT-SLAM~2.0 require substantially more GPU memory, reaching up to approximately 30\,GB in our experiments. Their high computational and memory demands, together with implementations that are not optimized for long trajectories, make them impractical for real-time operation on edge devices. Substantial model and memory optimization is therefore needed before these heavier learned methods can support long-duration edge deployment.

\section{Conclusions}

Large-area nadir SLAM is limited less by isolated tracking failures than by the absence of reliable global constraints: trajectories can remain continuous while progressively distorting beyond what a single alignment can absorb. In the two foundation-model systems analyzed, loop closure does not yet establish such constraints reliably: true revisits may be missed, while spatially inconsistent pairs may be accepted. Visual--inertial fusion can make metric scale observable and stabilize attitude, while altimetry and absolute localization against reference imagery can bound vertical and horizontal drift, respectively. Aerial-domain adaptation of foundation-model priors may improve local geometry, but dependable GNSS-denied navigation will likely require these complementary constraints.

\end{document}